\documentclass[conference]{IEEEtran}
\IEEEoverridecommandlockouts
\usepackage{cite}
\usepackage{amsmath,amssymb,amsfonts}
\usepackage{algorithmic}
\usepackage{graphicx}
\usepackage{textcomp}
\usepackage{xcolor}
\def\BibTeX{{\rm B\kern-.05em{\sc i\kern-.025em b}\kern-.08em
    T\kern-.1667em\lower.7ex\hbox{E}\kern-.125emX}}
\begin{document}

\title{3D Object Detection Method Based on YOLO and K-Means for Image and Point Clouds\\
{}
\thanks{}
}

\author{Xuanyu Yin$^{1,2}$ and Yoko Sasaki$^{2}$ and Weimin Wang$^{2}$ and Kentaro Shimizu$^{1}$% <-this % stops a space

\thanks{$^{1}$Graduate School of Interdisciplinary Information Studies, The University of Tokyo, The Tokyo, Japan
        {\tt\small yinxuanyu@g.ecc.u-tokyo.ac.jp}}%
\thanks{$^{2}$National Institute of Advanced Industrial Science and Technology, The Tokyo, Japan
        {\tt\small y-sasaki@aist.go.jp}}%
}

\maketitle

\begin{abstract}
Lidar based 3D object detection and classification tasks are essential for autonomous driving(AD). A lidar sensor can provide the 3D point cloud data reconstruction of the surrounding environment. However, real time detection in 3D point clouds still needs a strong algorithmic. This paper proposes a 3D object detection method based on point cloud and image which consists of there parts.(1)Lidar-camera calibration and undistorted image transformation. (2)YOLO-based detection and PointCloud extraction, (3)K-means based point cloud segmentation and detection experiment test and evaluation in depth image. In our research, camera can capture the image to make the Real-time 2D object detection by using YOLO, we transfer the bounding box to node whose function is making 3d object detection on point cloud data from Lidar. By comparing whether 2D coordinate transferred from the 3D point is in the object bounding box or not can achieve High-speed 3D object recognition function in GPU. The accuracy and precision get imporved after k-means clustering in point cloud. The speed of our detection method is a advantage faster than PointNet.
\end{abstract}


\section{Introduction}
Great progress has been made on 2D image understanding tasks, such as object detection and instance segmentation [1]. However, since the creation of 2D bounding boxes or pixel masks, real time detection on 3D point cloud data is becoming increasingly important in many applications areas, such as autonomous driving (AD) and augmented reality (AR). This paper presents our experiments on 3D object detection tasks, which are one of the most important tasks in 3D computer vision. Also presented is the analysis of the experimental results in precision, accuracy, recall, and time and possible future work to improve the 3D object detection average precision (AP).

In the AD field, the LIDAR sensor is the most common 3D sensor. It generates 3D point clouds and captures the 3D structure of scenes. The difficulty of point cloud-based 3D object detection mainly lies in the irregularity of the point clouds from LIDAR sensors  2]. Thus, state-of-art 3D object detection methods either leverage a mature 2D detection framework by projecting the point clouds into a bird's eye view or into a frontal view [2]. However, the information on the point cloud will suffer loss during the quantization process. Charles et al. at Stanford University published a paper on CVPR in 2017, in which he proposed a deep learning network called PointNet that directly handles point clouds. This paper was a milestone, marking the point cloud processing entered a new stage. The reason is that before PointNet, we had no way to deal with point clouds directly. Because point clouds are three-dimensional, they are not smooth. Moreover, deep neural networks, which make many ordinary algorithms, do not work. Thus, researchers have come up with a variety of methods[1][2][3], such as flattening the point clouds into pictures (MVCNN), dividing the point clouds into voxels, and then dividing them into nodes and straightening them in order. Thus, the cloud domain has advanced from the ``pre-PointNet era" to the ``post-PointNet era" thanks to the development of this technology. After PointNet, PointCNN, SO-Net, etc., came out, the operation of these methods improved steadily.

PointNet[3] has achieved 83.7 percent mean accuracy. However, the speed is still a problem. Compared with two-dimensional data, point cloud data with an additional dimension are too large to achieve the requirements for real-time 3D object detection. This paper presents our extraction of every point that may be an object after transformation in a 2D bounding box, enabling high-speed 3D object detection to be achieved. First, we describe a device we constructed including six cameras and one LIDAR. Then, we present the experiments we conducted to show how 2D images are captured by cameras and how 3D point clouds using LIDAR store the data in a rosbag, which was reused in subsequent experiments. The image data needed to be distorted, but an undistorted transform process was also needed. After the undistorted transform process, five images are split and dropped into the you only look once (YOLO) detection process. The YOLO detection process returns the related bounding box and class label. We store the bounding box and the class label for later reading, reducing the coupling of project research. In the second step, a function is written to extract the different topic information of the rosbag. After extracting the point cloud file, we put it into the numpy matrix for future operations.

Data conversion based on external and internal parameters is performed for every point cloud by matching the 2D images corresponding to each point cloud by matching the fps of the cameras and LIDAR. For each different bounding box of each point cloud, we collect all the matching points and render different colors based on different class labels. Finally, unsupervised clustering of point clouds in different bounding boxes improves the detection performance by removing some of the noise. The results of 3D object recognition are presented at the end of the paper. The recognition results were saved in a rosbag, and then 3D visualization was performed to check the experimental results. 

The evaluation results of the recognition experiment were completed using the method of depth images; the point cloud detection results were transferred to 32*1024 depth images. The final evaluation experiment was done for a comparison with ground truth in every pixel. The aforementioned is the rough research process of this article.

\section{3D object detection method}

\subsection{Overview}
\begin{figure}[h]
\centerline{\includegraphics[height=75mm]{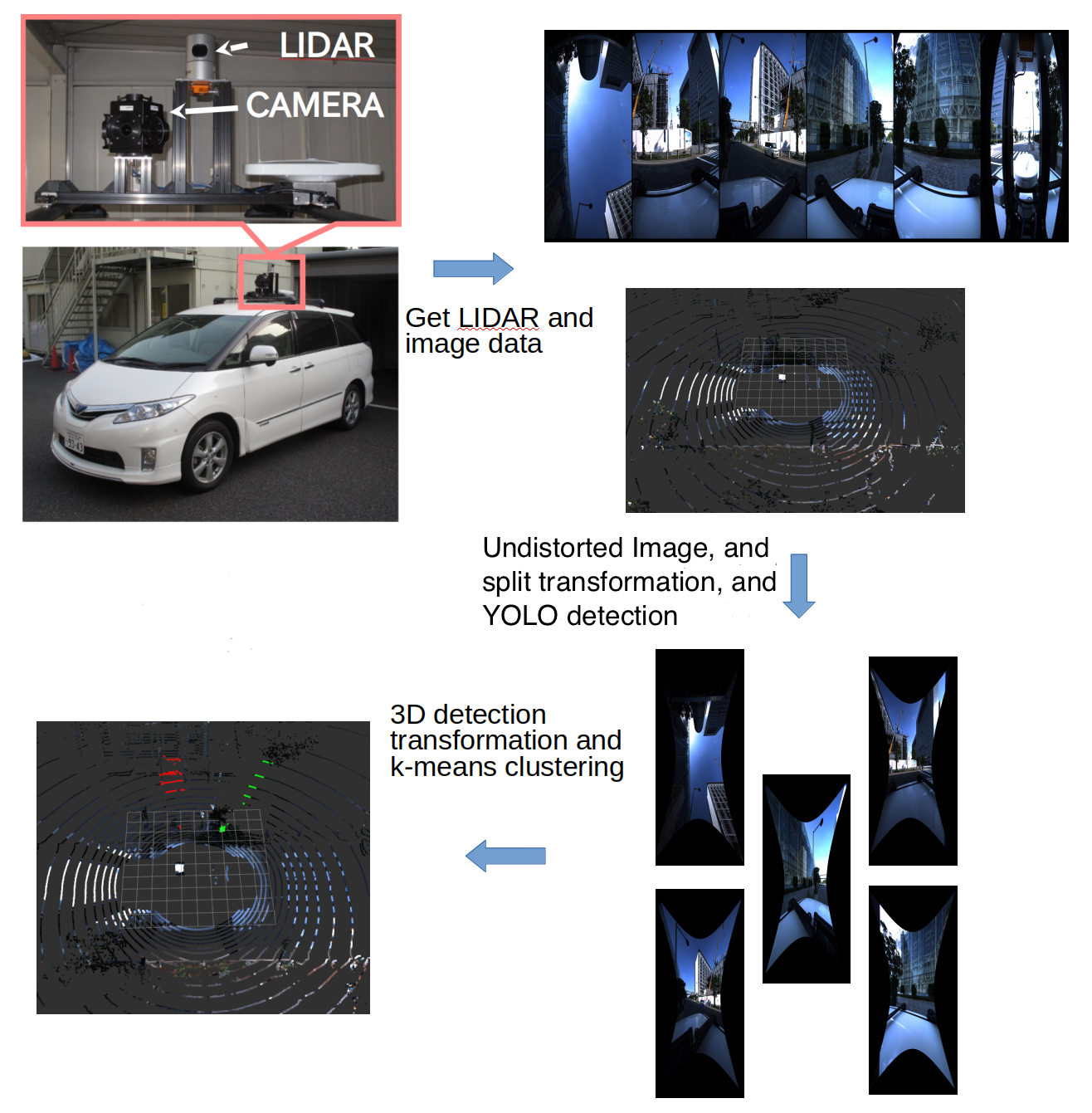}}
\caption{Overview of the proposed object detection system}
\label{fig}
\end{figure}
Fig.1 shows the overview of the proposed system. This research was basically divided into six parts. The first part mainly focused on the calibration of the cameras and the structural design of the testing equipment. The second part was to convert the distorted images into undistorted ones. The third part was YOLOv3 object recognition with 2D images. We mainly applied YOLOv1 tiny and YOLOv3 methods in doing the experiments, using keras to reproduce YOLO. The fourth part was the extraction of point clouds. We used rosbag to store the data and RVIZ for point cloud visualization. The fifth part was the unsupervised clustering of k-means, which were used to optimize the detection results of the basic experiments and to improve the detection accuracy of 3D object recognition.
\subsection{Lidar-camera calibration}
Here is the main information on the equipment used in this experiment and the external reference of the cameras.This experiment used Velodyne lidar(HDL-32e) with omni-directional cameras(PointGrey Ladybag5) to achieve 360 no dead angle monitoring, which is shown in Fig. 2.

\begin{figure}[h]
\centerline{\includegraphics[height=80mm]{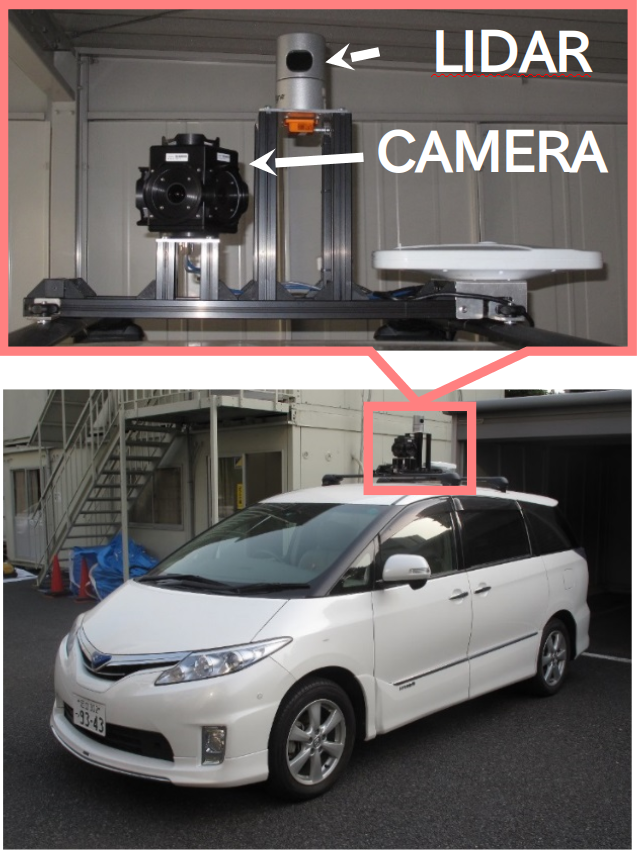}}
\caption{Structure of detection device: LIDAR and omni-directional camera}
\label{fig}
\end{figure}
A geometric model of camera imaging must be established during the image measurement process and machine vision application to determine the relationship between the three-dimensional geometric position of a point on the surface of a space object and its corresponding point in the image. These geometric model parameters are camera parameters. Under most conditions, these parameters must be obtained through experiments and calculations[4]. This process of solving parameters is called camera calibration. The internal parameters of the five cameras obtained at the end of this study are shown in Fig. 3.

\begin{figure}[h]
\centering{\includegraphics[height=25mm]{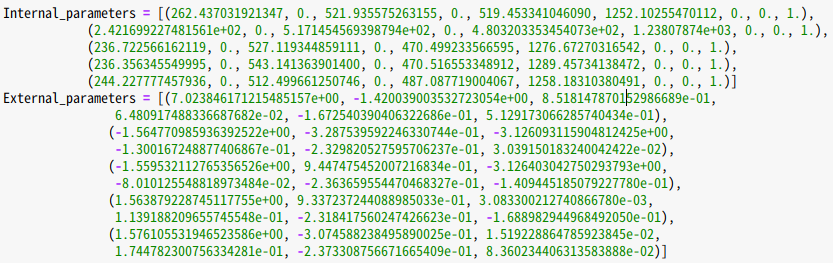}}
\caption{Internal and external parameters}
\label{fig}
\end{figure}
\subsection{Image undistorted transform}

In photography, wide-angle lenses are generally believed to be prone to barrel distortion, while telephoto lenses are prone to pincushion distortion. If a camera uses a short focal length wide-angle lens, the resulting image will be more susceptible to barrel distortion because the magnification of the lens gradually decreases as the distance increases, causing the image pixels to surround the center point radially. Fig. 4 shows raw images, and Fig. 5 shows images after the undistorted transform using OpenCV for image correction and camera calibration.
\begin{figure}[h]
\centering{\includegraphics[height=30mm]{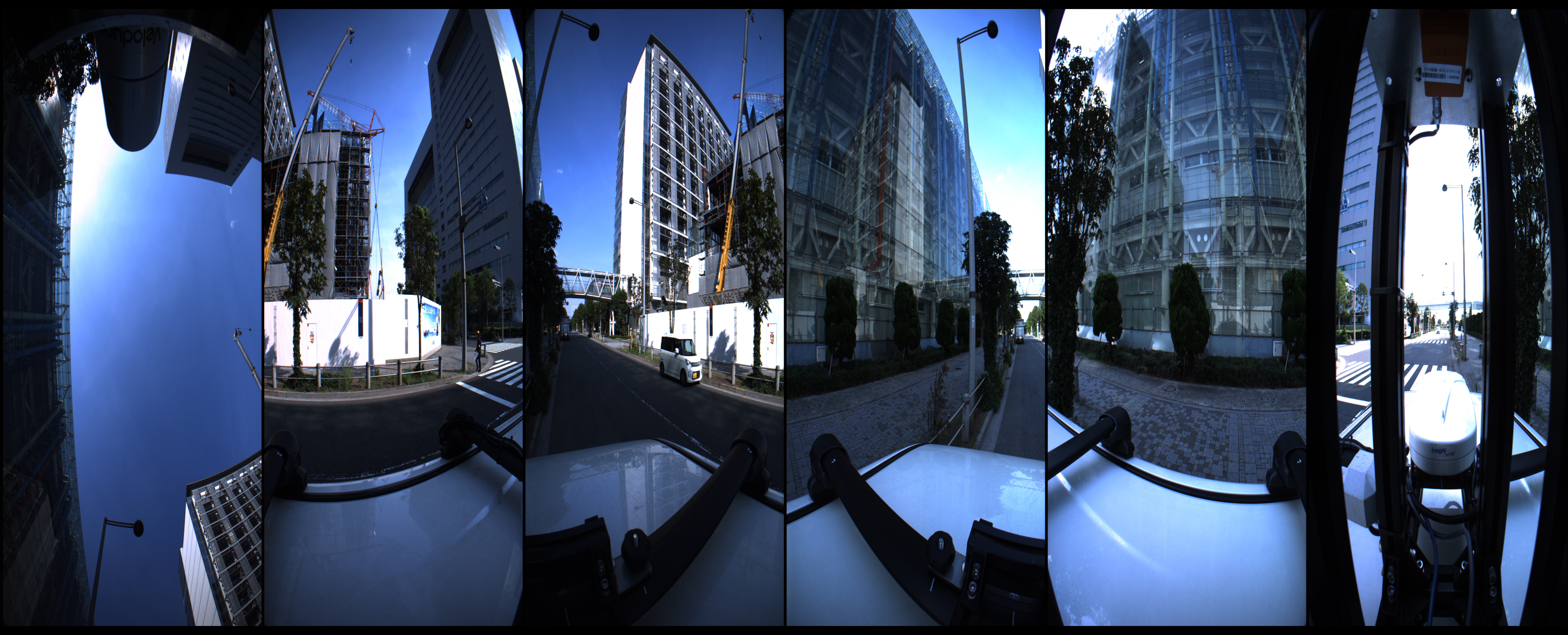}}
\caption{Raw images of five cameras}
\label{fig}
\end{figure}
\begin{figure}[h]
\centering{\includegraphics[height=30mm]{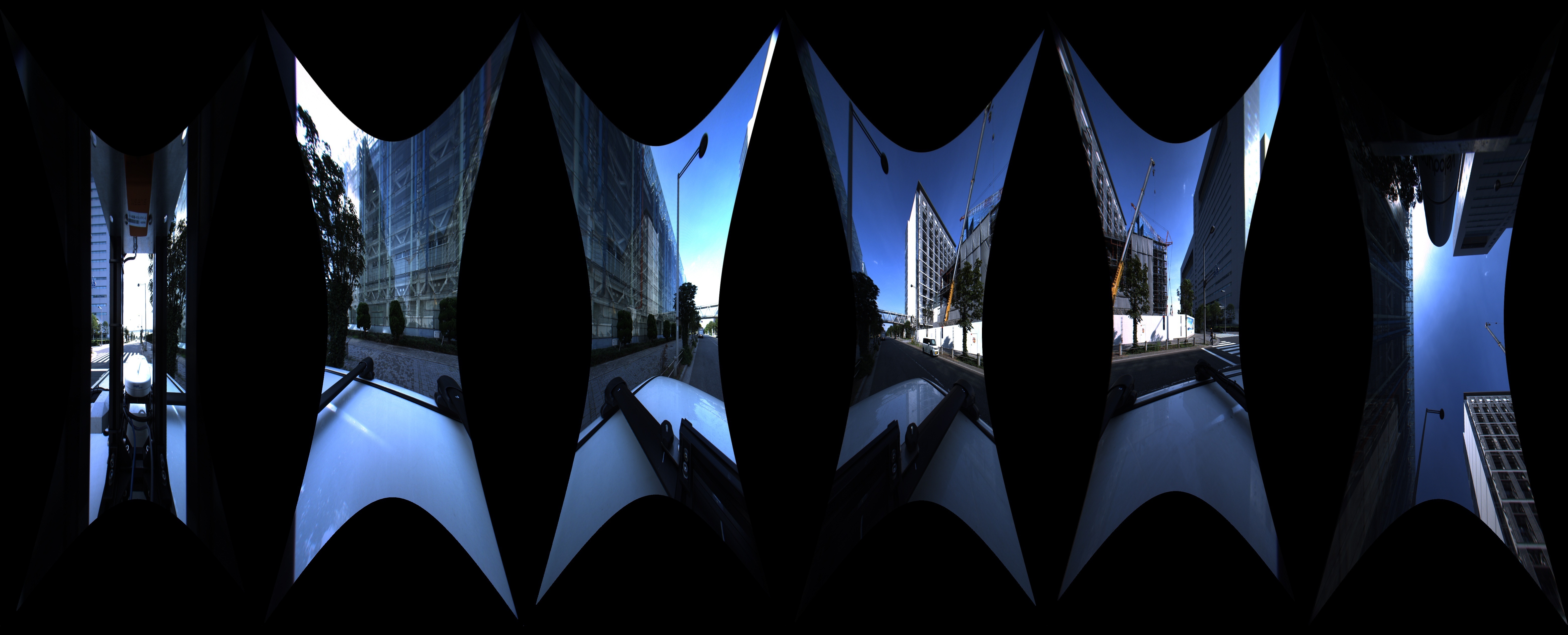}}
\caption{Images of five cameras after undistorted transform}
\label{fig}
\end{figure}
\subsection{YOLO-based detection}
YOLO is a fast target detection algorithm that is very useful for tasks with very high real-time requirements. The YOLO authors launched YOLOv3 version in 2018. After training on Titan X, v3 is 3.8 times faster than RetinaNet regarding mean average precision (mAP), and it can create a 320×320 picture in 22 ms. The objective score is 51.5, which is comparable to the accuracy of the single shot detector (SSD), but it is three times faster. Thus, YOLOv3 is very fast and accurate. In the case of IoU=0.5, it is equivalent to the mAP value of Focal Loss, but it is four times faster. We utilized YOLOv3 as a 2D object detection algorithm. Fig. 6 shows an example from a camera image, and Fig. 7 shows an example of 3D object detection from a point cloud.

We present a total of all the classes that can be identified in the coco dataset, including people, bicycles, cars, motorbikes, airplanes, buses, trains, trucks, boats, traffic, lights, fires, hydrants, and stop lights. In total, 80 classes are presented.

However, the most frequent classes are trucks, people, and cars. Thus, training a new YOLOv3 neural network to detect these three classes may be useful in saving detection time and enabling real-time functionality.

Because the images are largely black after undistorted conversion, we must remove the noise that exceeds its maximum bounding box when doing a conversion. Because the boundingbox of more than a quarter of the image size contains black parts, this part is invariably noisy. Fig 6 shows the YOLO detection examples from the camera image.
\begin{figure}[h]
\centering{\includegraphics[height=40mm]{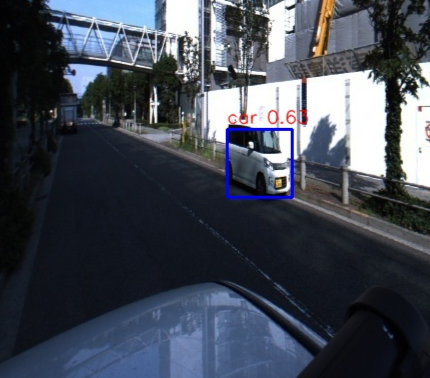}}
\caption{YOLO detection example from camera image}
\label{fig}
\end{figure}
\begin{figure}[h]
\centering{\includegraphics[height=45mm]{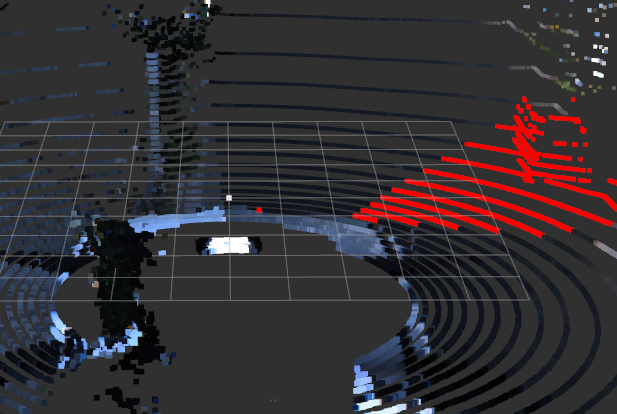}}
\caption{3D object detection example of point cloud}
\label{fig}
\end{figure}
\subsection{Point cloud extraction}
We mainly used rosbag to read and output the data. The read data contained undistorted images and point cloud images. The output was mainly the result of point cloud detection.
\subsection{K-means based point cloud segmentation}
\begin{figure}[h]
\centering{\includegraphics[height=35mm]{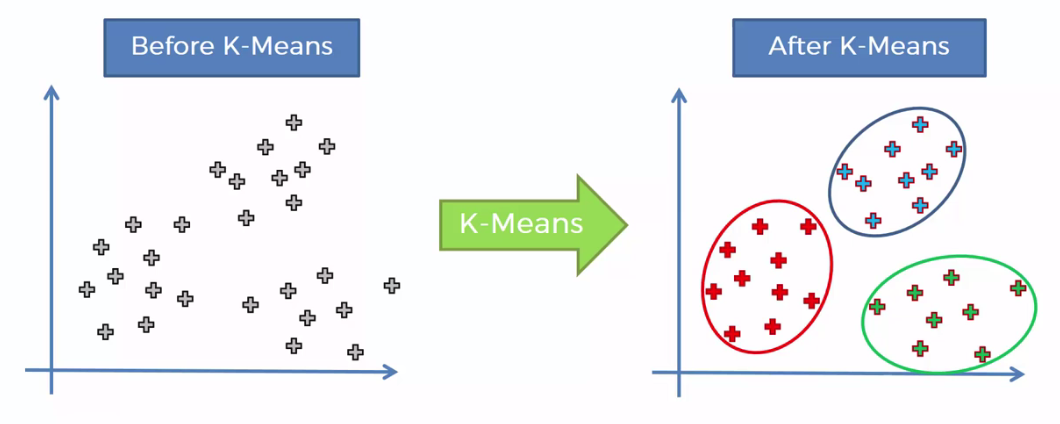}}
\caption{K-means clustering}
\label{fig}
\end{figure}

Because 2D is converted into 3D, all points that can be mapped into the bounding box in a certain direction are marked with different label colors. To improve the experiment, we utilized the unsupervised clustering method for k-means machine learning. The detection was faster, and the accuracy of the points substantially improved, removing most of the noise points. The k-means clustering graph is shown in Fig. 8, and the 3D object detection examples by this method is shown in Fig. 7.
\subsection{Evaluation of prediction results in depth images}
To create a comparison with ground truth and to make the results easier to observe, we converted the point clouds into a 3*1024 panoramic depth image that is shown in Fig. 9 with different colors corresponding to different categories. Depth image generation properties are shown in Table I. In this step, handmade 0.1K ground truth images were created using the LabelMe annotation tool. The final evaluation results also included these 0.1K pictures. 
\begin{figure}[h]
\centering{\includegraphics[height=10mm, width=90mm]{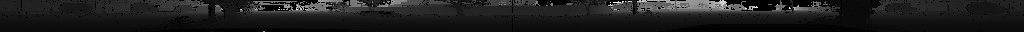}}
\caption{32*1024 depth image}
\label{fig}
\end{figure}
\begin{table}[htbp]
\caption{Depth image generation properties}
\begin{center}
\begin{tabular}{|c|c|}
\hline
Description&Specifications\\
\hline
Number of  detector pairs & 32\\
\hline
Limitation in  horizontal scanning&1024\\
\hline
Vertical  scanning  range& +10.67 to -30.67° \\
\hline
Angular resolution  in vertically& 1.33° \\
\hline
Angular resolution  in horizontally& 0.2°\\
\hline
Model  located area& 0 to 25,000[mm] from origin\\
\hline
Depth image  angle& 0 to 360 degrees \\
\hline
Output  depth image size& 32 × 1024 × 1\\
\hline
\end{tabular}
\label{tab1}
\end{center}
\end{table}
\section{Experiments and Results}
An in-vehicle sensor was used, and a large number of data were collected. The identification experiment consisted of two parts. The first included visualization and quantity statistics of point cloud identification with and without k-means clustering. The experimental results were then evaluated. The second part included evaluation criteria—the accuracy, precision, and recall—by comparing the results after conversion to depth images with ground truth.
\subsection{Experiments with different classes}
In this study, we first made 3D prediction results that were directly converted. Basically, all the points on the image that could be mapped to the bounding box were recognized. This led to a lot of noise, making it impossible to identify the 3D bounding box, but enabled recognizing the 3D radar data. A specific category exists in a certain direction, thereby completing the rough 3D recognition function. A 2D YOLO example is shown in Fig. 10, and related experiment results with and without k-means are shown in Fig. 11.

The proofreading here is mainly for the naked eye. In accordance with the specific bounding box data and the data under the camera, judgments were made as to whether or not the recognition results were correct.
\begin{figure}[h]
\centering{\includegraphics[height=40mm]{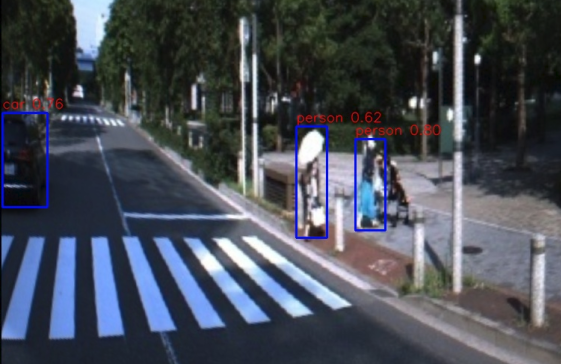}}
\caption{2D YOLO experimental results}
\label{fig}
\end{figure}
\begin{figure}[h]
\centering{\includegraphics[height=30mm]{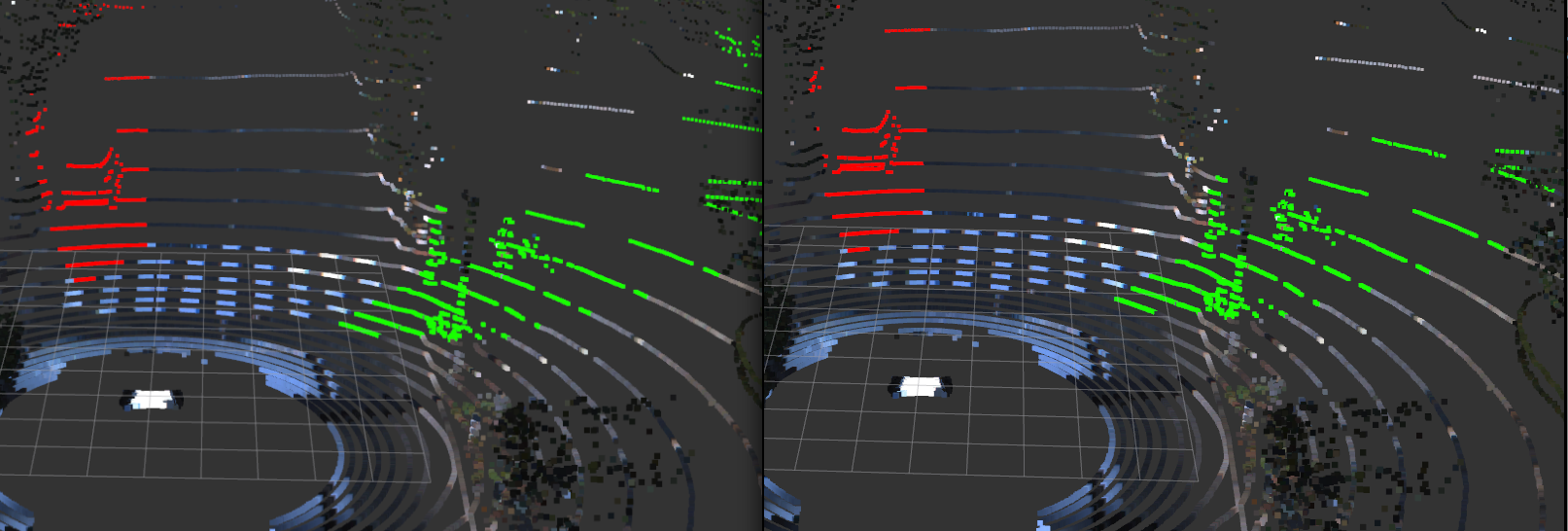}}
\caption{3D experiment results without(left) and with(right) k-means}
\label{fig}
\end{figure}
\subsection{Experiments with k-means}
K-means clustering is a method of vector quantization, originally from signal processing, which is popular for cluster analysis in data mining. K-means clustering aims to partition $n$ observations into $k$ clusters in which each observation belongs to the cluster with the nearest mean, serving as a prototype of the cluster. This results in a partitioning of the data space into Voronoi cells [4].

Because k-means pre-determines the number of categories and the maximum number of cluster iterations, the center point is randomly selected at first, so the results of each cluster become biased. However, the expected value of each category did not change too much at the end of the experiment. 

After using YOLO to complete the 2D object recognition and to convert the data into 3D point cloud data, we mainly used k-means to cluster further the points of the corresponding bounding box that had been acquired so as to remove some noise and to make the recognition results more accurate.

%\subsection{Experiments with Dbscan(if enough time)}
\subsection{Results with and without unsupervised learning}

The results of the experiment with and without k-means are shown in Fig. 11. The total number of point clouds in each group was 46,464. The maximum number of points was 4989. This occurred when a car or truck got very close. The lowest number was 0, which means that no target object could be recognized around that time. In other words, it was an empty space with no pre-trained class label objects.

Fig. 12 and Fig.13 represents the ratio and number of dropped data after clustering using the k-means method on the point cloud dataset. The highest was 49.15 percent in 46 frames. In the case of not recognizing any objects, the lowest was only 0.85 percent in 21 frames.

When clustering was not performed, the experimental results stained all the points mapped to the bounding box in that direction. After clustering, if there was an object, the number of point clouds in that part significantly increased, while other places that had no object obviously had a noticeable decrease in the number of clouds. Therefore, unsupervised clustering could significantly change the results, generally removing 31.6 percent of the point clouds.
\begin{figure}[h]
\centering{\includegraphics[height=45mm]{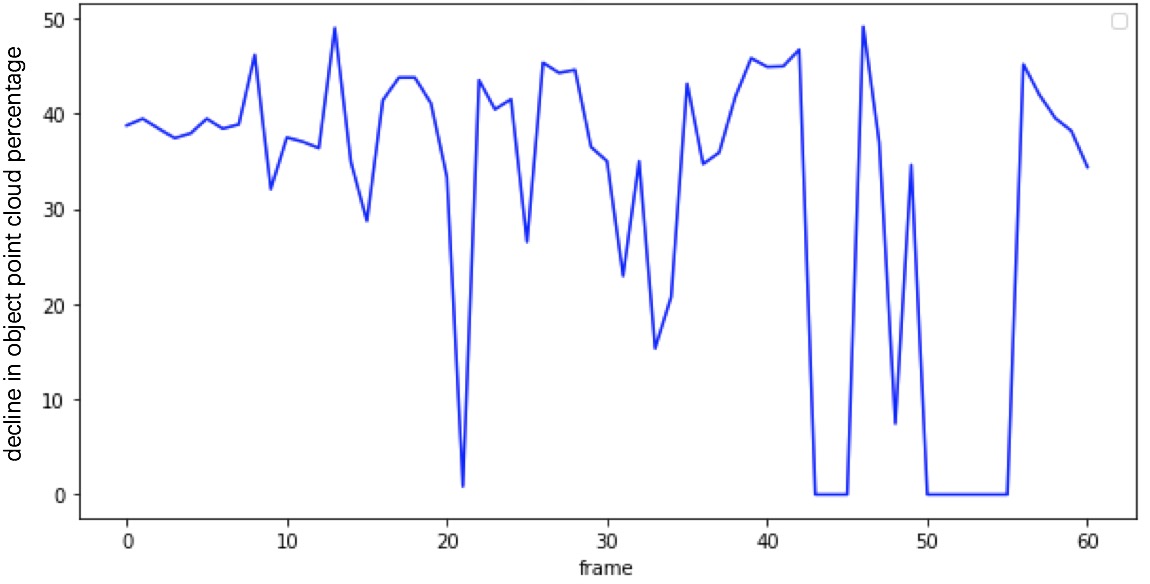}}
\caption{Rate of point cloud decline with k-means function}
\label{fig}
\end{figure}

\begin{figure}[h]
\centering{\includegraphics[height=45mm]{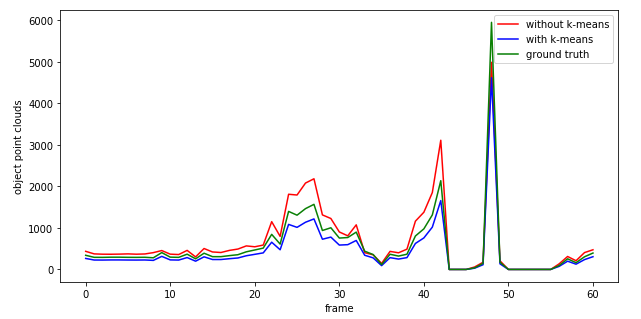}}
\caption{Experiment results of number of detection point clouds with and without k-means function}
\label{fig}
\end{figure}
\subsection{Evaluation of prediction results in depth images}
We made a 0.1K ground truth by labeling, reading, and outputting operations on LabelMe software\footnote{\noindent \textbf{labelme:}http://labelme.csail.mit.edu/Release3.0/}. In the previous stage, we got the prediction results of the point clouds. After converting the point clouds to the same 3*1024 depth image, the 3*1024*1 image file containing the object category labels were saved too. In the final evaluation test, accuracy, precision, and recall were calculated by comparing the ground truth and the numpy file holding the experimental prediction results. Fig. 14 is the Depth image which is converted from point cloud data. The ground truth is shown in Fig. 17, the detection results are in Fig. 16, and the final results after k-means clustering are in Fig. 17. The accuracy and consumed time results for the total prediction process are shown in Table II. Accuracy, precision, and recall with and without the k-means clustering were recorded. The results shown in Table III reveal YOLO 2D’s detection was successful.

\begin{table}[htbp]
\caption{Prediction results with and without k-means clustering}
\begin{center}
\begin{tabular}{|c|c|c|c|c|}
\hline
precision method&accuracy&time\\
\hline
PointNet &0.837 &1.00\\
\hline
image(YOLO)+pointcloud(without clustering) & 0.7288 &0.848\\
\hline
image(YOLO)+pointcloud(k-means clustering)&0.7301&0.856\\
\hline
\end{tabular}

\label{tab1}
The ``time" stand for the consumed time to handle one million point clouds (around 1K objects)
\end{center}
\end{table}
\begin{table}[htbp]
\caption{Prediction results after YOLO}
\begin{center}
\begin{tabular}{|c|c|c|c|c|}
\hline
precision method&accuracy&precision&recall\\
\hline
without clustering& 0.900 &0.7753 &0.6011\\
\hline
K-means clustering&0.9023&0.7867&0.6103\\
\hline
\end{tabular}
\label{tab1}
\end{center}
\end{table}
\begin{figure}[h]
\centering{\includegraphics[height=10mm, width=90mm]{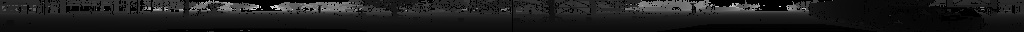}}
\caption{Depth image}
\label{fig}
\centering{\includegraphics[height=10mm, width=90mm]{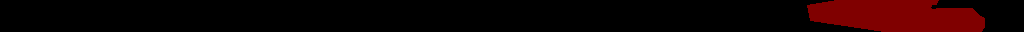}}
\caption{Ground truth}
\label{fig}
\centering{\includegraphics[height=10mm, width=90mm]{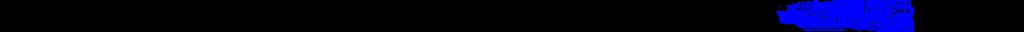}}
\caption{Prediction result without k-means}
\label{fig}
\centering{\includegraphics[height=10mm, width=90mm]{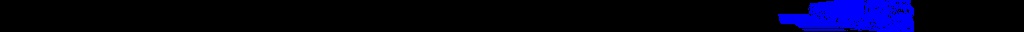}}
\caption{Prediction results with k-means function}
\label{fig}
\end{figure}

\section{Conclusions}
The conclusion of our study are as follows:

1.The method adopted by this paper is to directly convert the 3D point cloud to 2D image data, from the recognition of the 2D boudingbox to the dyeing of the 3D point cloud. Since the YOLO algorithm is adopted, the real-time performance is very strong, and the unsupervised clustering is used too. A lot of noise will be removed. It makes the recognition better.

2.This paper mainly wants to find a way to quickly and accurately determine whether there are objects and objects in a certain direction. This will contribute to the success of the unmanned field, allowing the car to obtain more information to make more judgments.

3.The final experimental results, in the case of using two 1080Ti GPUs, basically ensure that the experiment without clustering consumes 0.19 seconds per frame and 0.192 seconds after k-means clustering in 5 threads. The fast identification process ensures the real-time detection of the surrounding conditions in unmanned driving. If parallel, distributed computing and other technologies are used, the recognition speed will be faster.

4.The speed is very fast. However, the accuracy is not very high due to a front yolo recognition accuracy needs to be considered. The recall for detection is not high too.

\section{Future work}
In the future, robots will be added, and semantic mapping from running mobile robots will form the core of the next step. Then, we will consider not only the k-means function but also handling methods for directed point clouds like PointNet and FCN or more clustering methods like point cloud based depth clustering to figure out a faster method to complete 3d object detection using images and lidar. We will develop automatic labeling functions using our method for training data generation of LIDAR-based 3D objects.

\section*{Acknowledgment}

We would like to thank Wonjik Kim and Ryusei Hasegawa for providing us with their conversion tools, which transferred the point cloud data to 32*1024 depth images.

\vspace{12pt}
\color{red}


\begin{thebibliography}{00}


\bibitem{Shinjuku98}
Qi, C. Ruizhongtai, W. Liu, C. Wu, H. Su and L. J. Guibas. ``Frustum PointNets for 3D Object Detection from RGB-D Data,” 2018 IEEE/CVF Conference on Computer Vision and Pattern Recognition (2018): 918-927.
\bibitem{Shinjuku99}
S. Shaoshuai, W. Xiaogang and Li. Hongsheng ``PointRCNN: 3D Object Proposal Generation and Detection from Point Cloud,” CoRR abs/1812.04244 (2018): n. pag.


\bibitem{Shinjuku98}
Qi, C. Ruizhongtai, H. Su, K. Mo and L. J. Guibas. ``PointNet: Deep Learning on Point Sets for 3D Classification and Segmentation,” 2017 IEEE Conference on Computer Vision and Pattern Recognition (CVPR) (2017): 77-85.



\bibitem{Shinjuku98}
K. Jason, M. Mozifian, J. Lee, A. Harakeh and S. Lake Waslander. ``Joint 3D Proposal Generation and Object Detection from View Aggregation,” 2018 IEEE/RSJ International Conference on Intelligent Robots and Systems (IROS) (2018): 1-8.
\bibitem{b1} W. Kim, M. Tanaka, M. Okutomi, Y. Sasaki, ``Automatic Labeled LiDAR Data Generation based on Precise Human Model,'' in IEEE International Conference on Robotics and Automation (ICRA).IEEE, 2019.
\bibitem{Shinjuku99}
Mousaviaz, Arsalan, D. Anguelov, J. Flynn and J. Kosecka. ``3D Bounding Box Estimation Using Deep Learning and Geometry,” 2017 IEEE Conference on Computer Vision and Pattern Recognition (CVPR) (2017): 5632-5640.


\bibitem{Shinjuku98}
K. Shin, Y. Paul Kwon, M. Tomizuka, ``RoarNet:  A  Robust  3D  Object  Detection  based  onRegiOn  Approximation  Refinement,'' arXiv:1811.03818 [cs.CV].

\bibitem{Shinjuku99}
Y. Zhou, O. Tuzel, ``VoxelNet: End-to-End Learning for Point Cloud Based 3D Object Detection,'' 2017 IEEE Conference on Computer Vision and Pattern Recognition (CVPR)2017.

\bibitem{Shinjuku99}
Wikipedia, ``K-means clustering,'' https://en.wikipedia.org/\\wiki/K-means\_clustering
\bibitem{Shinjuku98}
C. Xiaozhi, M. Huimin, W. Ji, Li. Bo and X. Tian. ``Multi-view 3D Object Detection Network for Autonomous Driving,” 2017 IEEE Conference on Computer Vision and Pattern Recognition (CVPR) (2017): 6526-6534.


\bibitem{Shinjuku99}
Wikipedia, ``Tutorial Camera Calibration,'' https://boofcv.org\\/index.php?title=Tutorial\_Camera\_Calibration.
\bibitem{Shinjuku99}
C. Xiaozhi, K. Kundu, Z. Zhang, H. Ma, S. Fidler and R. Urtasun, ``Monocular 3D Object Detection for Autonomous Driving,” 2016 IEEE Conference on Computer Vision and Pattern Recognition (CVPR) (2016): 2147-2156.

\bibitem{b2} J. Sun, M. Ovsjanikov, and L. Guibas. ``A concise and provably informative multi-scale signature based on heat diffusion,'' In Computer graphics forum, volume 28, pages 1383–1392. Wiley Online Library, 2009.
\bibitem{b3} Z. Wu, S. Song, A. Khosla, F. Yu, L. Zhang, X. Tang, and J. Xiao. ``3d shapenets: A deep representation for volumetric shapes,'' In Proceedings of the IEEE Conference on Computer Vision and Pattern Recognition, pages 1912–1920, 2015.
\bibitem{b4} L. Yi, V. G. Kim, D. Ceylan, I.-C. Shen, M. Yan, H. Su, C. Lu, Q. Huang, A. Sheffer, and L. Guibas. ``A scalable active framework for region annotation in 3d shape collections,'' SIGGRAPH Asia, 2016.
\bibitem{b5} H. Ling and D. W. Jacobs. ``Shape classification using the inner-distance,'' IEEE transactions on pattern analysis and machine intelligence, 29(2):286–299, 2007
\bibitem{b6} M. Aubry, U. Schlickewei, and D. Cremers. ``The wave kernel signature: A quantum mechanical approach to shape analysis,'' In Computer Vision Workshops (ICCV Workshops), 2011 IEEE International Conference on, pages 1626–1633. IEEE, 2011.
\bibitem{b7} Y. Fang, J. Xie, G. Dai, M. Wang, F. Zhu, T. Xu, and E. Wong. ``3d deep shape descriptor,'' In Proceedings of the IEEE Conference on Computer Vision and Pattern Recognition, pages 2319–2328, 2015. 
\end{thebibliography}
\end{document}